\documentclass{article}
\usepackage{spconf,amsmath,graphicx}
\usepackage{epsfig}
\usepackage{amssymb}
\usepackage{mathrsfs}
\usepackage{import}
\usepackage{subcaption}
\usepackage{mathtools}
\usepackage{cancel}
\usepackage{xcolor}
\usepackage[utf8]{inputenc}
\usepackage[english]{babel}
\usepackage{amsthm}
\usepackage{appendix}
\usepackage{bbm}
\usepackage{cite}
\usepackage{theoremref}
\usepackage{hyperref}
\usepackage{balance}

\newtheorem{theorem}{Theorem}

\newcommand{\norm}[1]{\left\lVert#1\right\rVert}
\newcommand{\pushright}[1]{\ifmeasuring@#1\else\omit\hfill$\displaystyle#1$\fi\ignorespaces}

\title{Federated Learning with Local Differential Privacy: Trade-offs between Privacy, Utility, and Communication}
\name{Muah Kim$^{\star}$, Onur G{\"u}nl{\"u}$^{\star}$, and Rafael F. Schaefer$^{\dagger}$ \thanks{This work was supported in part by the German Federal Ministry of Education and Research (BMBF) within the national initiative for ``\emph{Post Shannon Communication (NewCom)}" under the Grant 16KIS1004 and in part by the German Research Foundation (DFG) under the Grant SCHA 1944/7-1.}}		
\address{\begin{tabular}{ccc}\normalsize 
		\begin{tabular}{c}
			$^{\star}$Information Theory and Applications Chair\\
			Technische Universit\"at Berlin\\
			Berlin, Germany\\[0.5ex]
			\texttt{\{muah.kim, guenlue\}@tu-berlin.de}
		\end{tabular}
		\begin{tabular}{c}
			$^{\dagger}$Lehrstuhl f\"ur Nachrichtentechnik/Kryptographie und Sicherheit\\
			Universit\"at Siegen\\
			Siegen, Germany\\[0.5ex]
			\texttt{rafael.schaefer@uni-siegen.de}
		\end{tabular}
\end{tabular}}

\begin{document}
\maketitle
\begin{abstract}
	Federated learning (FL) allows to train a massive amount of data privately due to its decentralized structure. Stochastic gradient descent (SGD) is commonly used for FL due to its good empirical performance, but sensitive user information can still be inferred from weight updates shared during FL iterations. We consider Gaussian mechanisms to preserve local differential privacy (LDP) of user data in the FL model with SGD. The trade-offs between user privacy, global utility, and transmission rate are proved by defining appropriate metrics for FL with LDP. Compared to existing results, the query sensitivity used in LDP is defined as a variable and a tighter privacy accounting method is applied. The proposed utility bound allows heterogeneous parameters over all users. Our bounds characterize how much utility decreases and transmission rate increases if a stronger privacy regime is targeted. Furthermore, given a target privacy level, our results guarantee a significantly larger utility and a smaller transmission rate as compared to existing privacy accounting methods.
\end{abstract}
\section{Introduction}
\label{sec:intro}

Differential privacy (DP) is widely used due to its strong privacy guarantees. DP tackles the privacy leakage about single data belonging to an individual in a dataset when some information from the dataset is publicly available. Common DP mechanisms add an independent random noise component to available data to provide privacy, which can be provided by using local sources of randomness such as physical unclonable functions (PUFs) \cite{gunlu2018key}. Applied to machine learning (ML), preserving DP of a training dataset is studied, e.g., in \cite{ duchi2014privacy, abadi2016deep,chaudhuri2011differentially, dwork2014algorithmic, shokri2015privacy}. Among various ML models, federated learning (FL) is a promising option due to its decentralized structure \cite{mcmahan2017communication,geyer2017differentially,wei2020federated}. Since user data are not collected by an aggregator in FL, local differential privacy (LDP) \cite{ding2017collecting} of an FL model is studied in the literature to guarantee individual user privacy, e.g., for a wireless multiple-access channel \cite{seif2020wireless}, by using splitting/shuffling \cite{sun2020ldp}, dimension selection \cite{liu2020fedsel}, with experimental evaluations \cite{truex2020ldp}, and with a communication-efficient algorithm \cite{Wang2020D2PFed}. Due to the iterative process in ML algorithms, the violation of LDP after multiple rounds of weight updates needs to be addressed. It can be done by using privacy accounting methods of DP such as the sequential composition theorem (SC) \cite{dwork2009differential}, the advanced composition theorem (AC1) \cite{dwork2010boosting, dwork2014algorithmic}, an improved advanced composition theorem (AC2) \cite{kairouz2015composition}, and the moments accountant (MA) \cite{abadi2016deep}. Unlike the composition theorems that directly compose the DP parameters, the MA approach circumvents the composition by converting DP into R\'{e}nyi-differential privacy (RDP), whose composition has a simple linear form. The MA is shown to outperform the AC1 and the AC2 when a Gaussian noise mechanism is used \cite{abadi2016deep}. The MA is further improved by using the optimal conversion from parameters of RDP to DP \cite{asoodeh2020better}. An FL model with stochastic gradient descent (FedSGD) with LDP is not yet investigated with a comprehensive theoretical analysis that considers privacy, utility, and communication jointly. This paper focuses on trade-offs between privacy, utility, and transmission rate, where LDP is provided to the FedSGD model by using a Gaussian mechanism.  In \cite{seif2020wireless}, the trade-offs between those three metrics are analyzed for the non-stochastic gradient descent algorithm for learning. Most of related studies consider the SC \cite{sun2020ldp, liu2020fedsel,truex2020ldp}, the AC1 \cite{seif2020wireless}, and the MA \cite{Wang2020D2PFed} for privacy accounting, all of which can be improved by using the privacy accounting method proposed in \cite{asoodeh2020better}. 

One main contribution of this paper is the privacy analysis for the FedSGD model by using an enhanced MA suggested in \cite{asoodeh2020better}. Furthermore, we propose a generic utility metric that considers the query sensitivity as a varying parameter, unlike in the literature, and we provide a lower bound on the utility metric. Our utility bound considers system heterogeneity by allowing users to have distinct dataset sizes, data sampling probabilities, and target privacy levels. The transmission rate is considered as the differential entropy of the noisy gradients for lossless communications. We illustrate significant gains from our bounds in terms of the required noise power, the utility metric, and the transmission rate as compared to the existing methods.

\section{System Model}
\label{sec:format}
\subsection{Federated SGD (FedSGD)}
\label{ssec:FedSGD}
Consider that the FedSGD method consists of a central server and $K$ users. The $K$ users are assumed to have their own neural networks with the same structure. At each time step $t$ such that $t\in\{1,2,\dots, T\}$, the server distributes the aggregated weight values $\mathbf{w}^{(t)}\in \mathbb{R}^d$ to all users and the $K$ users' networks are initialized with those weight values. Then, user $k$ randomly samples a dataset $\mathcal{J}_k^{(t)}$ from its whole dataset $\mathcal{D}_k$ with probability $q_k$ and calculates the gradient $\mathbf{g}_k^{(t)}$ from $\mathcal{J}_k^{(t)}$. In particular, we use a loss function $\ell(\mathbf{w}^{(t)},x)$ defined for each data sample $x$ and given weights $\mathbf{w}^{(t)}$, and we represent the local (per user) loss $\mathcal{L}_k(\mathbf{w}^{(t)},\mathcal{J}_k^{(t)})\in \mathbb{R}$ as \vspace{-5pt}
\begin{align}
\mathcal{L}_k(\mathbf{w}^{(t)},\mathcal{J}_k^{(t)})
&\coloneqq\dfrac{1}{|\mathcal{J}_k^{(t)}|}\sum_{x\in\mathcal{J}_k^{(t)}}\ell(\mathbf{w}^{(t)},x)\label{def_local_loss}
\end{align}
where $|\mathcal{J}_k^{(t)}|$ denotes the size of a set $\mathcal{J}_k^{(t)}$. Suppose the difference between the true loss and an empirical loss can be made negligible by using a sufficiently large dataset.
Each user's local gradient $\mathbf{g}_{k}^{(t)}$ is then represented as \vspace{-5pt}
\begin{align}
\mathbf{g}^{(t)}_k(\mathcal{J}_k^{(t)})\!=\!\nabla_{\mathbf{w}}\mathcal{L}_k(\mathbf{w}^{(t)},\mathcal{J}^{(t)}_k)\!=\!\dfrac{1}{|\mathcal{J}_k^{(t)}|}\!\sum_{x\in\mathcal{J}_k^{(t)}}\nabla_{\mathbf{w}}\ell(\mathbf{w}^{(t)},x)\nonumber
\end{align}
where $\nabla_{\mathbf{w}}$ represents the gradient along the weight vector $\mathbf{w}=(w_1,w_2,\dots,w_d)$, i.e.,$\nabla_{\mathbf{w}}=(\frac{\partial }{\partial w_1},\frac{\partial }{\partial w_2},\dots,\frac{\partial}{\partial w_d})$. Let $\norm{\mathbf{x}}$ denote the $\ell_2$-norm of a $d$-dimensional vector $\mathbf{x}=(x_1,x_2,\dots,x_d)$, i.e., $\norm{\mathbf{x}}=\sqrt{\sum_{i=1}^d x_i^2}$. Assume that $G$ is the maximum $\ell_2$-norm value of all possible gradients for any given weight vector $\mathbf{w}_k$ and sampled dataset $\mathcal{J}_k$, i.e.,  
$G=\sup_{\mathbf{w}_k\in\mathbb{R}^d,\mathcal{J}_k\in\mathcal{D}_k} \mathbb{E}[\norm{\mathbf{g}_k(\mathcal{J}_k)}]$. Each user clips its local gradient by a clipping threshold value $C\in(0,G]$ as
\begin{align}
\bar{\mathbf{g}}_k^{(t)} &= \mathbf{g}^{(t)}_k\Big/\max\Big\{1,\norm{\mathbf{g}^{(t)}_k}/C\Big\}.  
\end{align}
The clipped gradients $\{\bar{\mathbf{g}}_k^{(t)}\}_{k=1}^{K}$ are sent to the server and aggregated to obtain the updated weight vector $\mathbf{w}^{(t+1)}$. Suppose we use a learning rate $\eta_t$, then the weight update is 
\begin{align}
\mathbf{w}^{(t+1)} = \mathbf{w}^{(t)} -\eta_t\cdot \sum_{k=1}^{K}\dfrac{|\mathcal{J}_k^{(t)}|}{|\mathcal{J}^{(t)}|}\bar{\mathbf{g}}^{(t)}_k\label{eq_SGD}
\end{align}
where $\mathcal{J}^{(t)}=\cup_{k=1}^{K}\mathcal{J}_k^{(t)}$. 
The global loss of the FL system is 
\begin{align}
\mathcal{L}(\mathbf{w}^{(t)},\mathcal{J}^{(t)}) &=\sum_{x\in\mathcal{J}^{(t)}}\frac{1}{|\mathcal{J}^{(t)}|}\ell(\mathbf{w}^{(t)},x)\nonumber\\
&=\sum_{k=1}^{K} \frac{|\mathcal{J}_k^{(t)}|}{|\mathcal{J}^{(t)}|}\mathcal{L}_k(\mathbf{w}^{(t)},\mathcal{J}_k^{(t)})
\end{align}
which can be obtained by taking a weighted sum of the local losses defined in \eqref{def_local_loss}. Thus, the weighted average of local gradients is equivalent to the gradient of the global loss calculated with the whole sampled data, i.e., we have\vspace{-5pt}
\begin{align}
\nabla_{\mathbf{w}}\mathcal{L}(\mathbf{w}^{(t)},\mathcal{J}^{(t)})\nonumber&=\dfrac{1}{|\mathcal{J}^{(t)}|}\sum_{x\in\mathcal{J}^{(t)}}\nabla_{\mathbf{w}}\ell(\mathbf{w}^{(t)},x)\nonumber\\
&=\sum_{k=1}^{K}\dfrac{|\mathcal{J}_k^{(t)}|}{|\mathcal{J}^{(t)}|}\sum_{x\in\mathcal{J}_k^{(t)}}\dfrac{1}{|\mathcal{J}_k^{(t)}|}\nabla_{\mathbf{w}}\ell(\mathbf{w}^{(t)},x)\nonumber\\
&=\sum_{k=1}^{K}\dfrac{|\mathcal{J}_k^{(t)}|}{|\mathcal{J}^{(t)}|}\mathbf{g}^{(t)}_k(\mathbf{w}^{(t)},\mathcal{J}_k^{(t)}).
\end{align}
This implies that such an FL model results in the same weight update as the centralized model, so one global loss optimization problem can be divided into multiple local problems \cite{mcmahan2017communication}. 
\subsection{Local differential privacy (LDP)}
FL guarantees a certain level of privacy since the users do not directly send their data to the central server publicly \cite{mcmahan2017communication}. However, a certain amount of information can still be inferred from the shared information of the local networks, so a privacy mechanism is still necessary to protect user data. We consider LDP to guarantee individual user privacy. A mechanism $\mathcal{M}_k$ is $(\epsilon_k, \delta_k)-$LDP w.r.t. the user $k$'s dataset $\mathcal{D}_k$, if any two neighboring datasets $D\sim D'\subseteq\mathcal{D}_k$ satisfy for any $\mathcal{S}\subseteq\text{range}(\mathcal{M}_k)$ that \vspace{-5pt}
\begin{align}
&\Pr\left[\mathcal{M}_{k,D}(\mathbf{g}_k)\in\mathcal{S}\right] \nonumber\\
&\qquad\qquad\leq e^{\epsilon_k}\cdot\Pr\left[\mathcal{M}_{k,D'}(\mathbf{g}_k)\in\mathcal{S}\right] + \delta_k.
\end{align} 
We assume that the local gradients $\mathbf{g}^{(t)}_k$ are clipped and then LDP is satisfied by adding a Gaussian noise component $\mathbf{Z}_k$. Suppose the Gaussian noise variance of each dimension is proportional to $C^2$, i.e., $\mathbf{Z}_k\sim\mathcal{N}(\mathbf{0},C^2\sigma_k^2\mathbf{I}_d)$ for some $\sigma_k^2 >0$, where $\mathbf{I}_d$ is the $d\times d$ identity matrix. Denote the noisy gradients as $\tilde{\mathbf{g}}_k^{(t)}$, so the Gaussian LDP mechanism can be represented as $\tilde{\mathbf{g}}_k^{(t)} =\mathcal{M}_k(\bar{\mathbf{g}}^{(t)}_k) = \bar{\mathbf{g}}^{(t)}_k + \mathbf{Z}_k\sim \mathcal{N}(\bar{\mathbf{g}}^{(t)}_k ,C^2\sigma_k^2\mathbf{I}_d)$.
With such an LDP mechanism, the weight update equation \eqref{eq_SGD} can be used by replacing $\bar{\mathbf{g}}_k^{(t)}$ with $\tilde{\mathbf{g}}_k^{(t)}$ for all $t=1,2,\dots, T$.

\section{Trade-offs between Privacy, Utility, and Transmission Rate}\label{sec:pagestyle}
We next characterize the Gaussian noise variance required to guarantee a target LDP level after $T$ rounds of weight updates for FL with LDP, i.e., \emph{$T$-fold composition}. Furthermore, we analyze the effects of the Gaussian noise on the utility and the transmission rate, given a target LDP level. With an example we illustrate that a tighter privacy composition bound can yield a significantly larger utility and smaller transmission rate for the same target LDP level.
\subsection{Theoretical Analysis}
We define utility $\mathcal{U}(T)$ after $T$ iterations as the multiplicative inverse of the convergence rate, i.e., we have\vspace{-5pt}
\begin{equation}
\mathcal{U}(T)=\dfrac{1}{	\mathbb{E}[\mathcal{L}(\mathbf{w}^{(T)},\mathcal{J}^{(T)})]-\mathcal{L}(\mathbf{w}^*)}
\end{equation}
where $\mathbf{w}^*$ is the optimal weight vector that minimizes the global loss, i.e., $\mathbf{w}^*={\arg\min}_{\mathbf{w}\in\mathbb{R}^d}\mathcal{L}(\mathbf{w},\cup_{k=1}^K \mathcal{D}_k)$. This utility metric is used instead of accuracy to track the learning performance analytically. Consider that the transmission of a noisy gradient is lossless, and define the user $k$'s transmission rate $R_{\text{tr},k}$ by the differential entropy of its noisy gradient, i.e., $R_{\text{tr},k}\!=\!h(\tilde{\mathbf{g}}^{(t)}_k)$ for $t\!=\!1,2,\dots,T$. The transmission rate is measured by a differential entropy term for simplicity, which can be extended by allowing distortion. We remark that the transmission rate can be further reduced with quantization, as suggested, e.g., in \cite{shlezinger2020uveqfed}.

The following theorem provides the trade-offs between LDP parameters, utility, and the transmission rate for the assumed FL model; see Appendix~\ref{app:Theorem1proof} for its proof.\vspace{-3pt}

\begin{theorem}\label{THM:TRADEOFF}
	User $k$'s Gaussian mechanism $\mathcal{M}_k$, where $k=1,2,\ldots,K$, with noise variance of each dimension $C^2\sigma_k^2$ is $(\epsilon_k,\delta_k)$-LDP after $T$ rounds of weight updates for 
	\begin{align}
	&\epsilon_k>2\log(\delta_k^{-1})\max(\delta_k, \frac{1}{\sigma_k^2\ln(\frac{1}{q_k\sigma_k})}),\\
	&q_k<\frac{1}{16\sigma_k}, \qquad\text{ and }\qquad \sigma_k\geq1
	\end{align}
	if we have
	\vspace{-2pt}
	\begin{align}
	&\sigma_k^2 \!\geq\!  \dfrac{4q_k^2T}{1-q_k}\!\biggl[\! \dfrac{2}{\epsilon_k^2}\log{\dfrac{1}{\delta_k}} \!+\! \dfrac{1}{\epsilon_k} \!-\! \dfrac{2}{\epsilon_k^2}\biggl(\!\log(2\log\delta_k^{-1})\!+\! 1 \!-\! \log\epsilon_k\!\biggr)\!\!\biggr] \nonumber\\
	&\qquad\quad+\mathcal{O}\left(\dfrac{\log^2(\log\delta_k^{-1})}{\log\delta_k^{-1}}\right). \label{eq_noise}\vspace{-2pt}
	\end{align}
	\vspace*{-2pt}
	For a $\mu$-smooth and $\lambda$-strongly convex loss $\mathcal{L}(\mathbf{w},\mathcal{S})$ with respect to a $d$-dimensional weight vector $\mathbf{w}\in\mathbb{R}^d$ given an arbitrary subset $\mathcal{S}$ of $\mathcal{D}$ such that $\mathcal{S}\subseteq \mathcal{D}$ and for a learning rate $\eta_t=\frac{G}{C\lambda t}$, the utility of the noisy FedSGD model after $T$ iterations is bounded as
	\vspace{-3pt}
	\begin{align}
	\vspace{-2pt}\mathcal{U}(T)& \geq \frac{\lambda^2T}{\mu G^2}\min\left\{\frac{1}{2}, \frac{1}{1+d\sigma^2}\right\} \label{eq_utility}
	\end{align}
	where $\sigma^2=\frac{\sum_{k=1}^{K} (|D_k|q_k\sigma_k)^2}{(\sum_{k=1}^{K} |D_k|q_k)^2}$, and $G$ is the maximum value of the gradient. The transmission rate $R_{\text{tr},k}$ of user $k$ with the noise power of each dimension $C^2\sigma_k^2$ can be bounded as
	\begin{align}
	R_{\text{tr},k} \leq d\log(2\pi eC^2\sigma_k/\sqrt{d}).\label{eq_tr_rate_bound}
	\end{align}
\end{theorem}\vspace{-5pt}
\begin{proof}[Proof Sketch]
	The noise variance bound (\ref{eq_noise}) follows by extending \cite[Theorem 5]{asoodeh2020better}, where a generalized RDP-DP conversion is applied, to allow random sampling from each dataset with probability $q_k$ and a query sensitivity of $2C$ by using the proof of \cite[Lemma~3]{abadi2016deep}. The utility bound (\ref{eq_utility}) is obtained by extending \cite[Lemma 1]{rakhlin2011making} with an additional assumption of noisy SGD algorithm on the FL model. In particular, we introduce an adaptive learning rate $\eta_t$ that depends on the clipping threshold $C$ to bound the utility when the gradients are aggregated from $K$ users, where the noisy gradient of each user is obtained by randomly sampling the data, clipping the gradients, and adding Gaussian noise to the clipped gradients. It is followed by the transmission rate bound derived from the upper bound on the differential entropy when the random vector's covariance is upper-bounded.
\end{proof}

Theorem 1 illustrates the trade-offs between three metrics. If, e.g., the noise variance $\sigma^2$ is increased to guarantee stronger privacy, the utility is lower bounded by a smaller value and the transmission rate upper bound increases. For a sufficiently small $\delta_k$, the term in big $\mathcal{O}$ notation in \eqref{eq_noise} becomes negligible as compared to the other terms. In that case, the noise variance $\sigma^2$ is increasing linearly with $T$. With this choice of noise variance $\sigma^2\propto T$, the denominator of the utility bound \eqref{eq_utility} increases linearly with $T$. Thus, the utility bound converges to a constant even if $T$ tends to infinity. This implies that with a Gaussian noise mechanism used for LDP, the utility lower bound can be finite even when $T$ tends to infinity, and the achievement of the minimum loss $\mathcal{L}(\mathbf{w}^*)$ is not guaranteed. For this case, the maximum difference between the achieved loss and the optimal loss, which is not necessarily zero, is upper bounded.

\subsection{Local Privacy Accounting Method Comparisons}\label{subsec_sim}
We compare the bounds in Theorem \ref{THM:TRADEOFF} with the results obtained from the MA, AC1, and AC2. We do not consider the SC for comparison since the AC1 is known to outperform the SC method. Table~\ref{Table_noise_bound} lists the noise variance bounds of those composition methods when they are applied to the assumed model with a data sampling probability $q_k$ and a clipping threshold $C>0$, i.e., the query sensitivity is $2C$.

\begin{table}[t]
	\caption{Noise variance lower bounds used for comparisons.}
	\vspace*{-0.3cm}
	\begin{tabular}{|c | c |} 
		\hline
		Composition & Lower Bound of \\ [0.5ex] 
		Method & $\sigma_k^2$ \\
		\hline\hline
		Proposed &  \eqref{eq_noise} \\ 
		\hline
		MA \cite{abadi2016deep} &$\displaystyle  \frac{4q_k^2T}{1-q_k}\!\Bigl( \frac{2}{\epsilon_k^2}\log{\frac{1}{\delta_k}} \!+\! \frac{1}{\epsilon_k}+\mathcal{O}(\log\delta_k^{-1})\Bigr)  $  \\
		\hline
		AC1 \cite{dwork2010boosting, dwork2014algorithmic} & \eqref{eq_noise_str1} \\
		\hline
		AC2 \cite{kairouz2015composition}& $\displaystyle\frac{4q_k^2}{1-q_k}\frac{8T\log(e+\frac{\epsilon_k}{\delta_k})}{\epsilon_k^2}$ \\
		\hline
	\end{tabular}	\label{Table_noise_bound}
	\vspace*{-0.3cm}
\end{table}

We obtain the noise bound with the AC1 by using its implicit solution. Consider that each weight update round is $(\epsilon_0,\delta_0)-$LDP that results in $(\epsilon_k,\delta_k)-$LDP after $T$ iterations, which satisfies $\epsilon_k=\sqrt{2T\ln(\tilde{\delta}^{-1})}\epsilon_0+T\epsilon_0(e^{\epsilon_0}-1)$ and $\delta_k=T\delta_0+\tilde{\delta}$ by \cite[Theorem 3.20]{dwork2014algorithmic}. We obtain $(\epsilon_0,\delta_0)$ from given $(\epsilon_k,\delta_k)$ by choosing $\tilde{\delta}=10^{-5}$ and plug it into the noise bound of a Gaussian mechanism from \cite[Theorem 3.22]{dwork2014algorithmic} as 
\begin{align}
\sigma_k^2\geq \frac{4q_k^2}{1-q_k}\frac{2}{\epsilon_0}\log\left(\frac{4}{5\delta_0}\right). \label{eq_noise_str1}
\end{align}
The factor $\frac{4q_k^2}{1-q_k}$ of the noise bounds results from using the query sensitivity of $2C$ and randomly sampled datasets with probability $q_k$. For evaluations, we assume a homogeneous system consisting of $K\!=\!100$ users with the same parameters $q_k, \epsilon_k,\delta_k$, and $\sigma_k$. The aggregated noise variance $\sigma^2$ is then obtained as $\sigma^2=\frac{1}{K}\sigma_k^2$. We choose the minimum possible $\sigma_k^2$ for each method and calculate the utility bound by plugging the noise variance $\sigma^2$ into \eqref{eq_utility} with $\delta_k\!=\!10^{-4}, q_k\!=\!10^{-3}, d=10^4, \mu\!=\!1, \lambda=1, C=1$, and $G\!=\!5$.

\begin{figure}[t]
	\centering
	\begin{subfigure}{88mm}
		\centering
		\begin{minipage}[t]{43mm}
			\centering
			\includegraphics[width=43 mm]{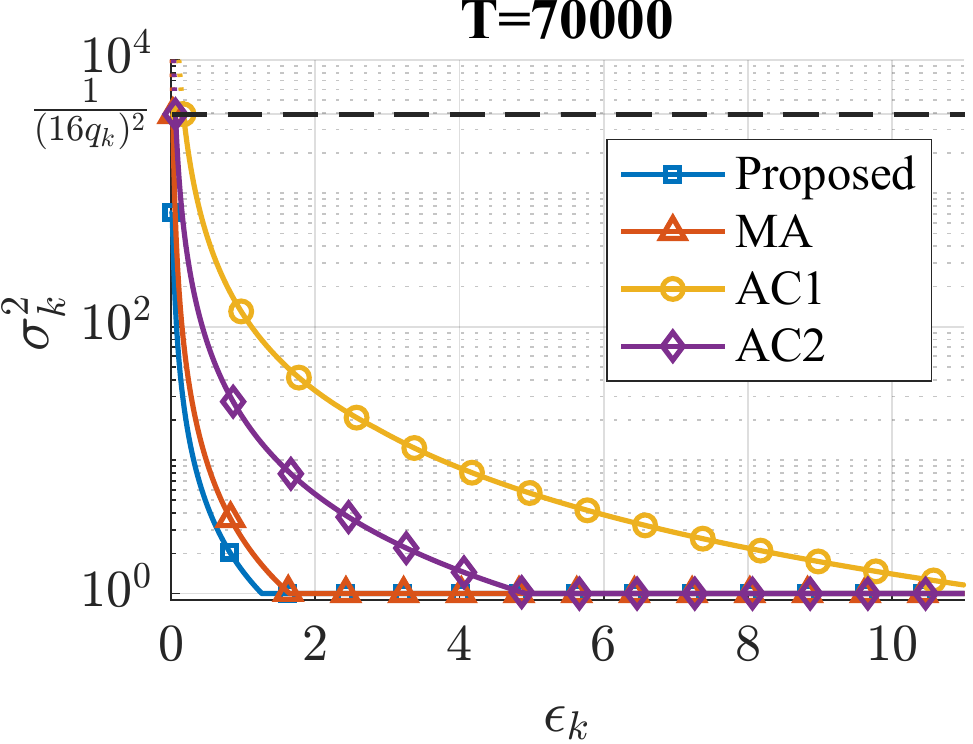}
		\end{minipage}
		\hfill
		\begin{minipage}[t]{43mm}
			\includegraphics[width=43 mm]{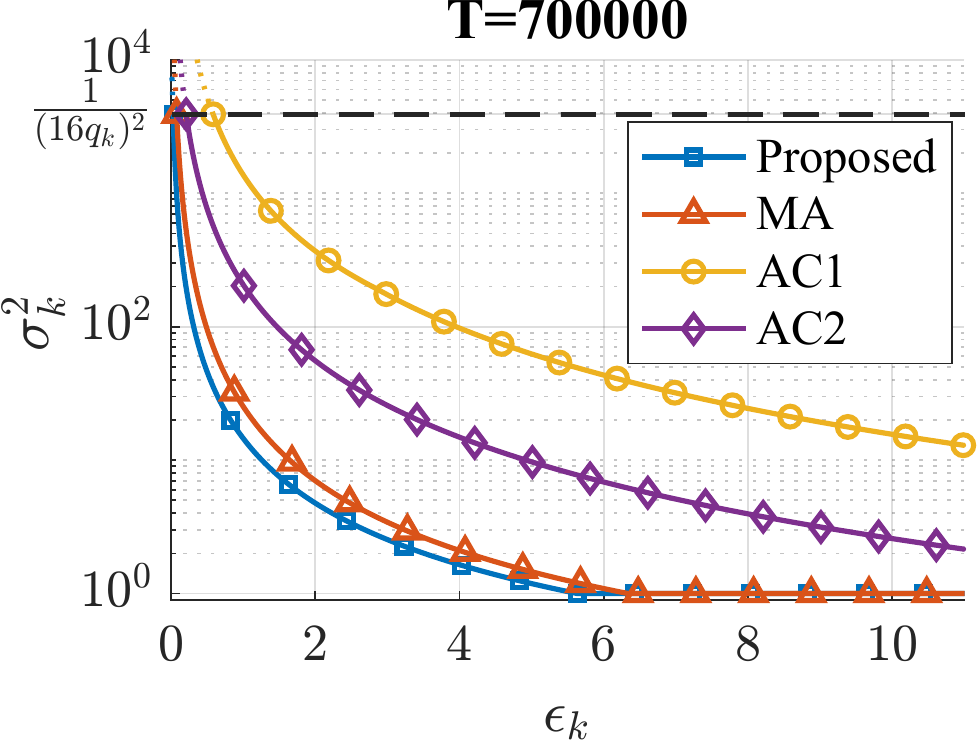}
		\end{minipage} 
		\vspace*{-0.2cm}
		\subcaption{$\epsilon_k$ vs. the lower bound on noise variance $\sigma_k^2$. }\label{Fig_noise}
	\end{subfigure}
	\begin{subfigure}{88mm}
		\centering
		\begin{minipage}[t]{43mm}
			\includegraphics[width=43 mm]{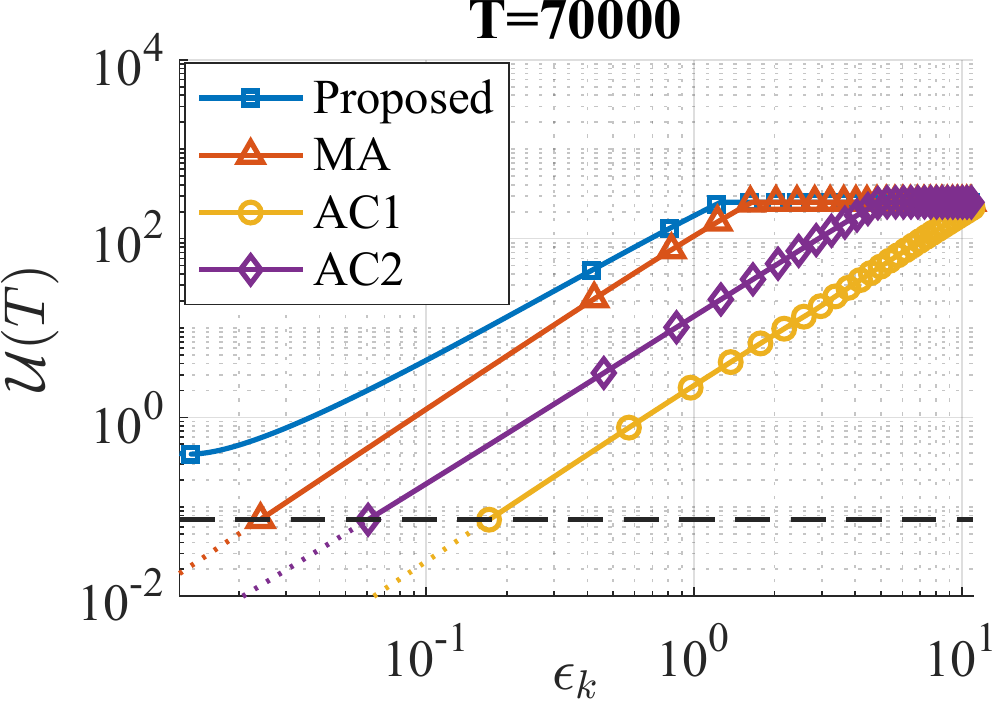}
		\end{minipage}
		\hfill
		\begin{minipage}[t]{43mm}
			\includegraphics[width=43 mm]{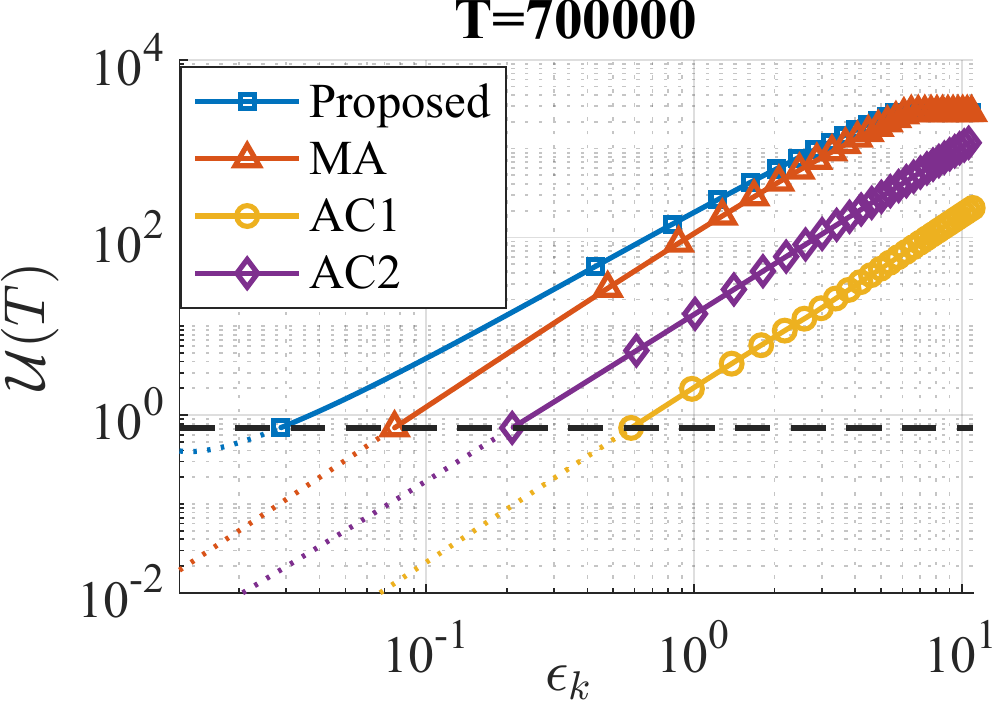}
		\end{minipage} 
		\vspace*{-0.15cm}
		\subcaption{$\epsilon_k$ vs. the lower bound on utility $\mathcal{U}(T)$.}\label{Fig_utility}
	\end{subfigure}
	\begin{subfigure}{88mm}
		\centering
		\begin{minipage}[t]{43mm}
			\includegraphics[width=43 mm]{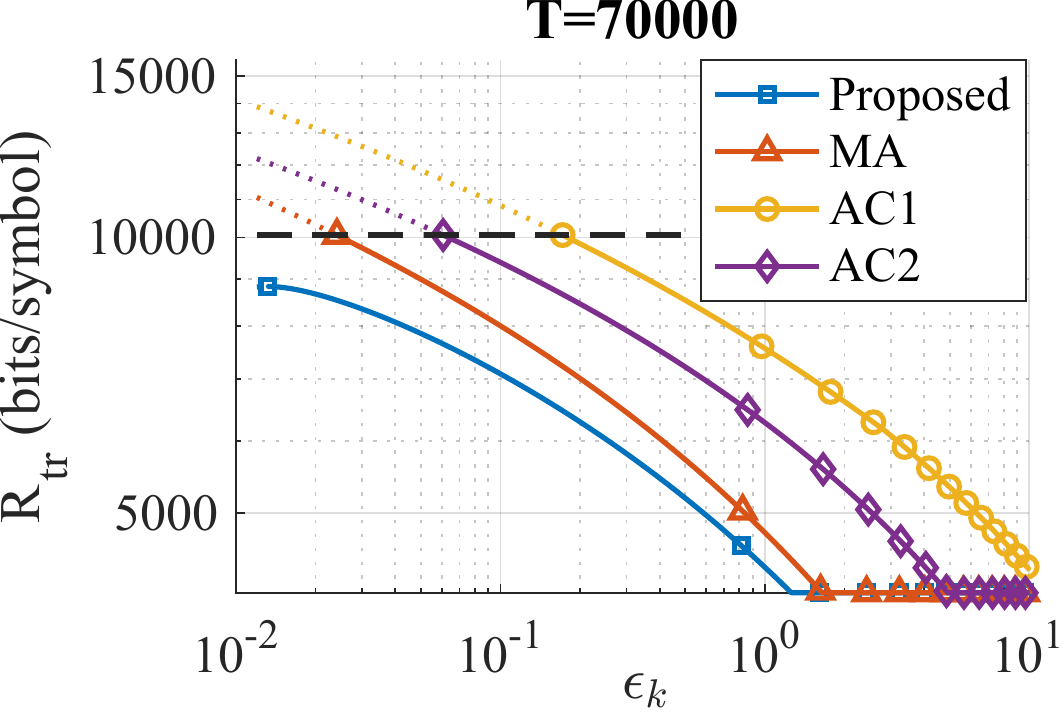}
		\end{minipage}
		\hfill
		\begin{minipage}[t]{43mm}
			\includegraphics[width=43 mm]{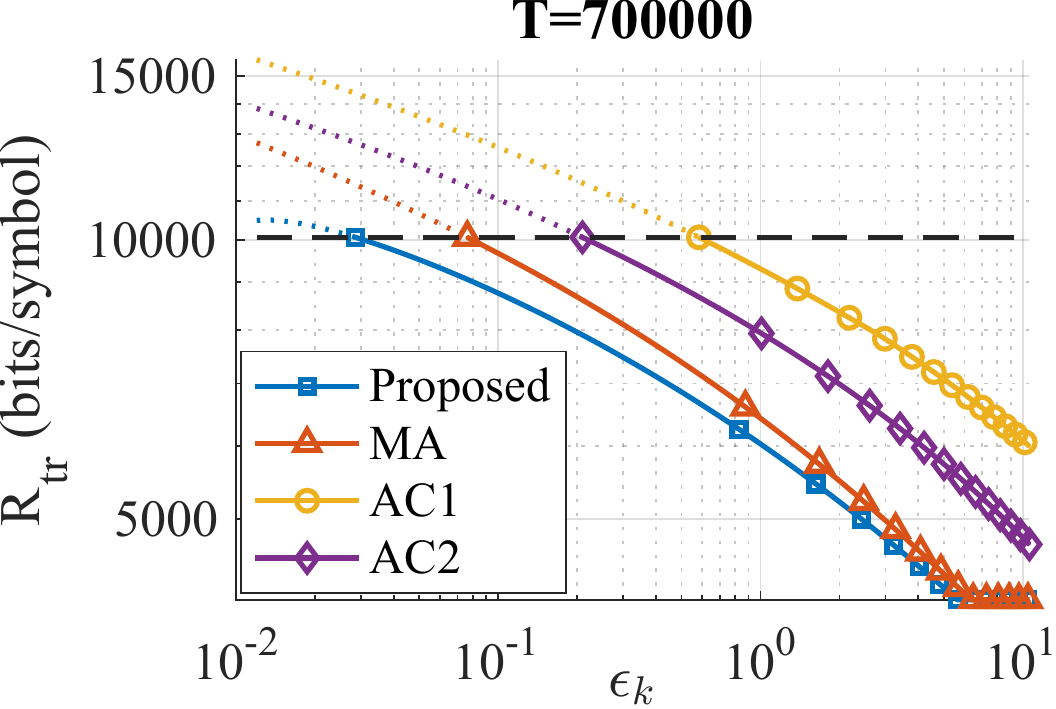}
		\end{minipage} 
		\vspace*{-0.15cm}
		\subcaption{$\epsilon_k$ vs. the upper bound on the transmission rate $R_{\text{tr},k}$.}\label{Fig_tr}
	\end{subfigure}
	\caption{The noise variance $\sigma_k^2$ and utility $\mathcal{U}(T)$ lower bounds vs. $\epsilon_k$ for $T\!=\!7\!\times\!10^4, 7\!\times\!10^5$ and with parameters $\delta_k\!=\!10^{-4}, q_k\!=\!10^{-3}$ for all $k\!=\!1,2,\dots,100, d\!=\!10^4, \mu\!=\!1, \lambda\!=\!1, C\!=\!1$, and $G\!=\!5$. }
	\label{Fig_sim}
	\vspace*{-0.5cm}
\end{figure}
The noise bound, the resulting utility, and the transmission rate are illustrated in Fig. \ref{Fig_sim} for parameters that satisfy corresponding constraints of each composition method for the number of iterations $T$ of $7\!\times\!10^4$ and $7\!\times\!10^5$. The obtained noise variance bounds are valid in the region such that $\sigma^2_k\!<\!1/(16q_k)^2\!=\!3906.25$, represented by the black dashed lines in Fig. \ref{Fig_noise}. This condition also limits the utility bounds at $\{0.0072, 0.0717\}$ for $T\!=\!\{7, 70\}\!\times\!10^4$, respectively, and the transmission rate bound at $9.69\!\times\!10^3$ (bits/symbol), which correspond to the black dashed lines in Fig. \ref{Fig_utility} and Fig. \ref{Fig_tr}, respectively. The noise bounds in Fig. \ref{Fig_noise} decrease to $1$ and then stay constant as the target privacy level $\epsilon_k$ increases. These conditions are from \cite[Lemma 3]{abadi2016deep}, which are required to measure the effect of randomly sampled data on the noise variance. The utility bound and the transmission rate bound might be improved if the noise variance bound does not require the condition $\sigma_k\!\geq\!1$. We remark that $\epsilon$ values shown in \cite[Fig.~3]{asoodeh2020better} seem to be unfortunately wrong; furthermore, \cite[Theorem~5]{asoodeh2020better} seems to lack a condition on $\epsilon$ that follows from \cite[Lemma 3]{abadi2016deep}, which imposes a condition on $\alpha$ used in \cite[Eq. (80)]{asoodeh2020better}. The violation of this condition makes the range of $T$ values considered in \cite[Fig.~3]{asoodeh2020better} invalid.

The proposed method requires the smallest noise variance, followed by MA, AC2, and AC1, for the same $\epsilon_k$ for every $T$ considered in Fig. \ref{Fig_noise}. The required noise variance of every method increases with $T$ because more iterations deteriorate individual privacy and a larger amount of noise is necessary to meet the same target privacy. Fig. \ref{Fig_utility} shows the utility bounds obtained with $\sigma_k^2$ values from Fig. \ref{Fig_noise}. The utility bound curves stay constant after they reach their maximum values. The maximum value of the utility bound increases with $T$ as the weight vector gradually converges to the optimum at every iteration even if the noise variance also increases. Thus, the utility bound is not degraded by the noise variance if the target privacy $\epsilon_k$ is large enough and the noise variance $\sigma^2_k$ is small enough for homogeneous users. If we target $\epsilon_k\!=\!0.3$ after $T\!=\!7\!\times\!10^4$ iterations, for example, the guaranteed utility bounds are at $25.83, 10.79, 0.22$, and $1.40$ when the proposed bound, MA, AC1, and AC2 are used, respectively. The corresponding transmission rate bounds are at $\{5.81, 6.44, 9.26, 7.91\}\times10^3$ (bits/symbol). Hence, a significantly larger utility bound and a smaller transmission rate can be achieved for the same privacy constraint $\epsilon_k$ with a tighter privacy composition bound.
\section{Conclusion}
Trade-offs between privacy, utility, and transmission rate of a FedSGD model with a Gaussian LDP mechanism were proved. We provided a noise variance bound that guarantees a given LDP level after multiple rounds of weight updates by using a tight composition theorem. The proposed utility bound allows distinct parameters for all users and allows the gradients to be clipped and noisy. The noise variance required is illustrated to be significantly smaller than the ones obtained by using existing privacy composition methods MA, AC1, and AC2. Similarly, our bounds lead to a significantly larger utility and a smaller transmission rate. In future work, we will illustrate gains from our bounds as compared to existing methods for large available datasets used for FL.

\bibliographystyle{IEEEbib}
\bibliography{refs}

\appendix
\begin{appendices}
	\section{Proof of Theorem~\ref{THM:TRADEOFF}}\label{app:Theorem1proof}
	Clipping gradient norms with $C$ makes the query sensitivity to be at most $2C$ because for arbitrary clipped gradients $\bar{\mathbf{g}}_1$ and $\bar{\mathbf{g}}_2$ the following inequality holds:
	\begin{equation*}
	\max_{\forall \bar{\mathbf{g}}_1,\bar{\mathbf{g}}_2 }\!\sqrt{\sum_{i\in[d]}(\bar{g}_{1,i}\!-\!\bar{g}_{2,i})^2} \!=\! \max_{\forall \bar{\mathbf{g}}_1 }\!\sqrt{\sum_{i\in[d]}(\bar{g}_{1,i}\!-\!(\!-\bar{g}_{1,i}))^2}\!\leq\! 2C.
	\end{equation*}
	The proof of \eqref{eq_noise} follows mainly from \cite[Theorem 5]{asoodeh2020better}, which provides the lower bound of noise variance when the query sensitivity is $1$ and the same dataset is repeatedly used at every iteration. \cite[Theorem 5]{asoodeh2020better} is obtained by using MA through the following steps:
	\begin{itemize}
		\item Compute $\gamma(\alpha)$ such that the Gaussian mechanism is $(\alpha,\gamma(\alpha))-$RDP for a given $\alpha$. 
		\item Apply the linear composition of RDP \cite{abadi2016deep}, i.e., the model is $(\alpha, T\gamma(\alpha))-$RDP after $T$ iterations. 
		\item Convert $(\alpha, T\gamma(\alpha))-$RDP into $(\epsilon(\alpha, \delta),\delta)-$DP for a given $\delta$.   
	\end{itemize}
	
	As compared to \cite[Theorem 5]{asoodeh2020better}, Theorem \ref{THM:TRADEOFF} considers a randomly sampled dataset due to the SGD and also considers a fixed but general query sensitivity $2C$. We first obtain the $\gamma_k(\alpha_k)$ parameter for the Gaussian mechanism of user $k$ for an arbitrary integer $\alpha_k$ when the dataset is obtained by applying random sampling with probability $q_k$. We then obtain the noise variance $\sigma_k^2$ bound by using MA. To avoid confusion, we denote the RDP cost of \cite[Theorem 5]{asoodeh2020better} by $\gamma_0(\alpha)=\frac{\alpha_k}{2\sigma_k^2}$ and that of Theorem \ref{THM:TRADEOFF} by $\gamma_k(\alpha_k)$. We obtain $\gamma_k(\alpha_k)$ by using the same method used in the proof of \cite[Lemma~3]{abadi2016deep}.
	
	Suppose we use a Gaussian mechanism $\mathcal{M}_k(\mathcal{D}_k)=\bar{\mathbf{g}}_k(\mathcal{D}_k)+\mathcal{N}(\mathbf{0},C^2\sigma_k^2\mathbf{I}_d)$ for a given dataset $\mathcal{D}_k$ and the corresponding clipped gradient $\bar{\mathbf{g}}_k$. Consider a neighboring dataset $\mathcal{D}_k'$ to $\mathcal{D}_k$ that only differs by a single data $D_n$, i.e., $\mathcal{D}_k=\mathcal{D}'_k\cup \{D_n\}$ without loss of generality. Instead of considering the worst case that $\bar{\mathbf{g}}_k(\mathcal{D}_k)=-\bar{\mathbf{g}}_k(\mathcal{D}'_k)$ and $\norm{\bar{\mathbf{g}}_k(\mathcal{D}_k)}=C$, we assume  $\bar{\mathbf{g}}_k(\mathcal{D}'_k)=\mathbf{0}$ and $\bar{\mathbf{g}}_k(\mathcal{D}_k)=2C\mathbf{e}_1$ and analyze the R\'{e}nyi divergence of two perturbed gradients without loss of generality. This assumption makes the problem one-dimensional because the neighboring datasets $\mathcal{D}_k$ and $\mathcal{D}'_k$ result in the gradients that have different elements only in the first dimension. Let $\mu_0$ denote the PDF of $\mathcal{N}(0,C^2\sigma_k^2)$ and $\mu_1$ denote the PDF of $\mathcal{N}(2C,C^2\sigma_k^2)$. 
	The noisy gradients of the Gaussian mechanism can be represented in one dimension as
	\begin{align}
	&\mathcal{M}_k(\mathcal{D}'_k) \sim \mu_0\\
	&\mathcal{M}_k(\mathcal{D}_k) \sim \mu \coloneqq (1-q_k)\mu_0 + q_k\mu_1.
	\end{align}
	To observe $\gamma_k$ for given $\mu$, $\mu_0$, and an integer $\alpha_k$, we start with the definition of RDP:
	\begin{align}
	&\dfrac{1}{\alpha_k}\mathbb{E}_{z\sim\mu}[(\mu(z)/\mu_0(z))^{\alpha_k}]\leq \gamma_k  \\
	&\dfrac{1}{\alpha_k}\mathbb{E}_{z\sim\mu_0}[(\mu_0(z)/\mu(z))^{\alpha_k}]\leq \gamma_k \label{ineq_gamma_def}.
	\end{align}
	Two inequalities can be shown by the same method, so we show only the second inequality here.
	Changing the probability that the expectation is taken over from $\mu_0$ to $\mu$ and using the binomial expansion, we have
	\begin{align}
	&\mathbb{E}_{z\sim\mu_0}[(\mu_0(z)/\mu(z))^{\alpha_k}] =\mathbb{E}_{z\sim\mu}[(\mu_0(z)/\mu(z))^{\alpha_k+1}] \nonumber\\
	&=\mathbb{E}_{z\sim\mu}[(1+(\mu_0(z)-\mu(z))/\mu(z))^{\alpha_k+1}]\nonumber\\
	&=\sum_{i=0}^{\alpha_k+1}\dbinom{\alpha_k+1}{i}\mathbb{E}_{z\sim\mu}[((\mu_0(z)-\mu(z))/\mu(z))^{i}].
	\end{align}
	The first term of the summation coming from $i=0$ simply becomes $1$. The second term when $i=1$ is $0$ by simple calculus. The third term with $i=2$ can be bounded as
	\begin{align}
	&\mathbb{E}_{z\sim\mu}[((\mu_0(z)-\mu(z))/\mu(z))^{2}]\nonumber\\
	&=\mathbb{E}_{z\sim\mu}[((q_k\mu_0(z)-q_k\mu_1(z))/\mu(z))^{2}]\nonumber\\
	&=q_k^2\int_{-\infty}^{\infty}(\mu_0(z)-\mu_1(z))^2/\mu(z)dz\nonumber\\
	&\leq\frac{q_k^2}{1-q_k}\int_{-\infty}^{\infty}(\mu_0(z)-\mu_1(z))^2/\mu_0(z)dz\nonumber\\
	&=\frac{q_k^2}{1-q_k}\mathbb{E}_{z\sim\mu_0}[((\mu_0(z)-\mu_1(z))/\mu_0(z))^{2}].
	\end{align}
	The expected value of the above term can be further simplified and bounded as below:
	\begin{align}
	&\mathbb{E}_{z\sim\mu_0}[((\mu_0(z)-\mu_1(z))/\mu_0(z))^{2}]\nonumber\\
	&=\mathbb{E}_{z\sim\mu_0}\left[\left(1-\exp\left(\frac{-z^2+4Cz-4C^2+z^2}{2C^2\sigma_k^2}\right)\right)^{2}\right]\nonumber\\
	&=1 \!-\! 2\mathbb{E}_{z\sim\mu_0}\left[\exp\left(\dfrac{4Cz \!-\! 4C^2}{2C^2\sigma_k^2}\right)\right] \nonumber\\
	&\qquad+\! \mathbb{E}_{z\sim\mu_0}\left[\exp\left(\dfrac{8Cz \!-\! 8C^2}{2C^2\sigma_k^2}\right)\right]\nonumber\\
	&=1 \!-\! 2 \!+\! \exp\left(\frac{8C^2}{2C^2\sigma_k^2}\right) \nonumber\\
	&= \exp\left(\frac{4}{\sigma_k^2}\right) \!-\! 1 \!=\! 4/\sigma_k^2 \!+\! \mathcal{O}\left(\frac{1}{\sigma_k^4}\right).
	\end{align}
	The third term with $i=2$ eventually becomes
	\begin{align}
	&\dbinom{\alpha_k+1}{2}\mathbb{E}_{z\sim\mu} \! \left[\left(\dfrac{\mu_0(z)-\mu(z)}{\mu(z)}\right)^2\right] \nonumber\\
	&\quad\leq \! \dfrac{4\alpha_k(\alpha_k+1)q_k^2}{2(1-q_k)\sigma_k^2}+\mathcal{O}\left(\frac{q_k^2\alpha_k^2}{\sigma_k^4}\right).
	\end{align}
	\cite[Lemma 3]{abadi2016deep} shows that the other terms, i.e., $(i\geq 3)$ terms, are upper bounded by $\mathcal{O}\left(\dfrac{q_k^3\alpha_k^3}{\sigma_k^3}\right)$. Thus, $\gamma_k(\alpha_k)=\dfrac{2 q_k^2}{(1-q_k)\sigma_k^2}(\alpha_k+1) + \mathcal{O}(\dfrac{q_k^3\alpha_k^2}{\sigma_k^3})$. Note that using $\gamma_k(\alpha_k)$ with $\sigma_k^2$ is equivalent to using $\gamma_0(\alpha_k)$ with $\frac{(1-q_k)\alpha_k}{4q_k^2(\alpha_k+1)}\sigma_k^2$ because
	\begin{align}
	\gamma_k(\alpha_k)=\frac{2q_k^2(\alpha_k+1)}{(1-q_k)\sigma_k^2}=\frac{\alpha_k}{2\left(\frac{(1-q_k)\alpha_k}{4q_k^2(\alpha_k+1)}\sigma_k^2\right)}.
	\end{align}
	In accordance with an assumption that $\alpha_k$ is optimally chosen to be $\alpha_k=2\log(\delta_k^{-1})/\epsilon_k$ from \cite[Theorem 5]{asoodeh2020better}, we obtain $\alpha_k/(\alpha_k+1)\approx 1$ when $\delta_k$ is sufficiently small. We can obtain the following inequality by directly using \cite[Theorem 5]{asoodeh2020better} for the noise variance associated with $\gamma_k(\alpha_k)$:
	\begin{align}
	&\frac{1-q_k}{4q_k^2}\sigma_k^2\approx \frac{(1-q_k)\alpha_k}{4q_k^2(\alpha_k+1)}\sigma_k^2\nonumber\\
	&\quad\!\geq\!\frac{2T}{\epsilon_k^2}\log\frac{1}{\delta_k} \!+\! \frac{T}{\epsilon_k} -\frac{2T}{\epsilon_k^2}\left(\log(2\log\delta_k^{-1}) \!+\! 1 \!-\! \log\epsilon_k\right)\nonumber\\ 
	&\hspace{+50pt}+\mathcal{O}\left(\frac{\log^2(\log\delta_k^{-1})} {\log\delta_k^{-1}}\right).
	\end{align}
	Dividing both hand sides by the coefficient of $\sigma_k^2$ completes the proof of \eqref{eq_noise}.
	
	Due to the assumption that $\mathcal{L}$ is $\lambda$-strongly convex, for all $\mathbf{w},\mathbf{w}'\in\mathbb{R}^d$ and any subgradient $\mathbf{g}$ of $\mathcal{L}$ at $\mathbf{w}$ we have
	\begin{equation}
	\mathcal{L}(\mathbf{w}')-\mathcal{L}(\mathbf{w})\geq \langle\mathbf{g},\mathbf{w}'-\mathbf{w}\rangle+\frac{\lambda}{2}\norm{\mathbf{w}'-\mathbf{w}}^2
	\end{equation}
	so that we obtain
	\begin{align}
	\langle\mathbf{g}^{(t)}_k ,&\mathbf{w}^{(t)}-\mathbf{w}^* \rangle = -\langle\mathbf{g}^{(t)}_k ,\mathbf{w}^*-\mathbf{w}^{(t)} \rangle \nonumber\\
	&\geq \mathcal{L}(\mathbf{w}^{(t)})-\mathcal{L}(\mathbf{w}^*)+\frac{\lambda}{2}\norm{\mathbf{w}^{(t)}-\mathbf{w}^*}^2\geq 0\label{ineq_convecity1}
	\end{align}
	and   
	\begin{align}	
	\mathcal{L}&(\mathbf{w}^{(t)})-\mathcal{L}(\mathbf{w}^*)\geq \langle\mathbf{g}^*,\mathbf{w}^{(t)}-\mathbf{w}^*\rangle+\frac{\lambda}{2}\norm{\mathbf{w}^{(t)}-\mathbf{w}^*}^2\nonumber\\
	&\hspace{6.5em}\geq\frac{\lambda}{2}\norm{\mathbf{w}^{(t)}-\mathbf{w}^*}^2\label{ineq_convecity2},
	\end{align}
	where $\mathbf{g}^*$ is the gradient at the optimum point $\mathbf{w}^*$ calculated with an arbitrary subset $\mathcal{S}$ of $\mathcal{D}$. The $\mu$-smoothness condition gives
	\begin{equation}
	\mathbb{E}[\mathcal{L}(\mathbf{w}^{(t)})-\mathcal{L}(\mathbf{w}^*)] \leq\mathbb{E}\left[\frac{\mu}{2}\norm{\mathbf{w}^{(t)}-\mathbf{w}^*}^2\right]\label{ineq_smooth}.
	\end{equation}
	
	We want to next observe how the weight vector $\mathbf{w}^{(t)}$ converges to its optimum $\mathbf{w}^*$ by bounding the expected mean square error $\mathbb{E}\left[\norm{\mathbf{w}^{(t)}-\mathbf{w}^*}^2\right]$. We first obtain a recurrence formula for $\norm{\mathbf{w}^{(t)}-\mathbf{w}^*}^2$ by using the weight update equation \eqref{eq_SGD} with $\bar{\mathbf{g}}_k^{(t)}$ replaced by $\tilde{\mathbf{g}}_k^{(t)}$ as follows: 
	\begin{align}
	&\norm{\mathbf{w}^{(t+1)}-\mathbf{w}^*}^2=\norm{\mathbf{w}^{(t)}-\mathbf{w}^*-\eta_t\cdot \sum_{k=1}^{K}\frac{|\mathcal{J}_k^{(t)}|}{|\mathcal{J}^{(t)}|}\tilde{\mathbf{g}}^{(t)}_k}^2\nonumber\\
	&=\norm{\mathbf{w}^{(t)}-\mathbf{w}^*}^2 \!+\! \eta_t^2\norm{\sum_{k=1}^{K}\frac{|\mathcal{J}_k^{(t)}|}{|\mathcal{J}^{(t)}|}\tilde{\mathbf{g}}^{(t)}_k}^2\nonumber
	\\
	&\qquad-2\eta_t\langle\mathbf{w}^{(t)}-\mathbf{w}^*,\sum_{k=1}^{K}\frac{|\mathcal{J}_k^{(t)}|}{|\mathcal{J}^{(t)}|}\tilde{\mathbf{g}}^{(t)}_k\rangle\label{eq_se_bound}.
	\end{align}
	First, the second term can be decomposed into
	\begin{align}
	\eta_t^2\norm{\sum_{k=1}^{K}\frac{|\mathcal{J}_k^{(t)}|}{|\mathcal{J}^{(t)}|}\tilde{\mathbf{g}}^{(t)}_k}^2&=\eta_t^2\norm{\sum_{k=1}^{K}\frac{|\mathcal{J}_k^{(t)}|}{|\mathcal{J}^{(t)}|}(\bar{\mathbf{g}}^{(t)}_k+\mathbf{Z}_k)}^2\nonumber\\
	&=\eta_t^2\norm{\sum_{k=1}^{K}\frac{|\mathcal{J}_k^{(t)}|}{|\mathcal{J}^{(t)}|}\bar{\mathbf{g}}^{(t)}_k+\mathbf{Z}}^2
	\end{align}
	where $\mathbf{Z}=\sum_{k=1}^{K}\frac{|\mathcal{J}_k^{(t)}|}{|\mathcal{J}^{(t)}|} \mathbf{Z}_k$. The square of the $\ell_2$-norm can be expanded as
	\begin{align*} 
	\eta_t^2\norm{\sum_{k=1}^{K}\frac{|\mathcal{J}_k^{(t)}|}{|\mathcal{J}^{(t)}|}\bar{\mathbf{g}}^{(t)}_k}^2+&2\eta_t^2\sum_{k=1}^{K}\frac{|\mathcal{J}_k^{(t)}|}{|\mathcal{J}^{(t)}|}\langle\bar{\mathbf{g}}^{(t)},\mathbf{Z}\rangle+
	\eta_t^2\norm{\mathbf{Z}}^2.
	\end{align*}
	It is straightforward to show that $\mathbf{Z}\sim\mathcal{N}(\mathbf{0},C^2\sigma^2\mathbf{I}_d)$ where $\sigma^2=\frac{\sum_{k=1}^{K} (|\mathcal{D}_k|q_k\sigma_k)^2}{(\sum_{k=1}^{K} |\mathcal{D}_k|q_k)^2}$. 
	The expected value of the second term can be obtained as follows. 
	\begin{align}
	&\mathbb{E}\left[\eta_t^2\norm{\sum_{k=1}^{K}\frac{|\mathcal{J}_k^{(t)}|}{|\mathcal{J}^{(t)}|}\bar{\mathbf{g}}^{(t)}_k}^2\right]+\mathbb{E}\left[\cancelto{0}{2\eta_t^2\sum_{k=1}^{K}\frac{|\mathcal{J}_k^{(t)}|}{|\mathcal{J}^{(t)}|}\langle\bar{\mathbf{g}},\mathbf{Z}\rangle}\right]\nonumber\\
	&\qquad\qquad+\mathbb{E}[
	\eta_t^2\norm{\mathbf{Z}}^2]\nonumber\\
	&\leq\eta_t^2C^2+d\eta_t^2C^2\sigma^2 = \eta_t^2C^2(1+d\sigma^2)\label{ineq_2nd_term_bound}
	\end{align}
	which follows because every element of $\mathbf{Z}$ is a zero-mean Gaussian, and taking the inner product is a linear transformation. Next, the third term of \eqref{eq_se_bound} can be decomposed into the inner products that can be locally calculated as
	\begin{align}
	&-2\eta_t\langle\mathbf{w}^{(t)}-\mathbf{w}^*,\sum_{k=1}^{K}\frac{|\mathcal{J}_k^{(t)}|}{|\mathcal{J}^{(t)}|}\tilde{\mathbf{g}}^{(t)}_k\rangle\nonumber \\
	&\qquad=-2\eta_t\sum_{k=1}^{K}\frac{|\mathcal{J}_k^{(t)}|}{|\mathcal{J}^{(t)}|}\langle\mathbf{w}^{(t)}-\mathbf{w}^*,\tilde{\mathbf{g}}^{(t)}_k\rangle.\label{eq_3rd_term}
	\end{align}
	We bound the expected value of each summand in (\ref{eq_3rd_term}) as
	\begin{align}
	&\mathbb{E}\left[\frac{|\mathcal{J}_k^{(t)}|}{|\mathcal{J}^{(t)}|}\langle\mathbf{w}^{(t)}-\mathbf{w}^*,\tilde{\mathbf{g}}^{(t)}_k\rangle\right] \nonumber\\
	&\!=\!\mathbb{E}\!\left[\!\frac{|\mathcal{J}_k^{(t)}|}{|\mathcal{J}^{(t)}|}\!\langle\mathbf{w}^{(t)} \!-\! \mathbf{w}^*,\bar{\mathbf{g}}^{(t)}_k\rangle\!\right] \! + \!\cancelto{0}{ \mathbb{E}\!\left[\!\frac{|\mathcal{J}_k^{(t)}|}{|\mathcal{J}^{(t)}|}\langle\!\mathbf{w}^{(t)} \!-\! \mathbf{w}^*,\mathbf{Z}_k\rangle\!\right]\!}\nonumber\\
	&\geq\mathbb{E}\left[\frac{|\mathcal{J}_k^{(t)}|}{|\mathcal{J}^{(t)}|}\frac{C}{G}\langle\mathbf{w}^{(t)}-\mathbf{w}^*,\mathbf{g}^{(t)}_k\rangle\right].
	\end{align}
	And then we bound the expected value of \eqref{eq_3rd_term} as follows:
	\begin{align}
	&-2\eta_t\mathbb{E}\left[\langle\mathbf{w}^{(t)}-\mathbf{w}^*,\sum_{k=1}^{K}\frac{|\mathcal{J}_k^{(t)}|}{|\mathcal{J}^{(t)}|}\tilde{\mathbf{g}}^{(t)}_k\rangle\right] \nonumber\\
	&\leq -2\eta_t\sum_{k=1}^{K}\mathbb{E}\left[\frac{|\mathcal{J}_k^{(t)}|}{|\mathcal{J}^{(t)}|}\frac{C}{G}\langle\mathbf{w}^{(t)}-\mathbf{w}^*,\mathbf{g}^{(t)}_k\rangle\right]\nonumber\\
	&\overset{(a)}{\leq} -2\eta_t\frac{C}{G}\mathbb{E}\left[\mathcal{L}(\mathbf{w}^{(t)})-\mathcal{L}(\mathbf{w}^*)+\frac{\lambda}{2}\norm{\mathbf{w}^{(t)}-\mathbf{w}^*}^2\right]  \nonumber\\
	&\overset{(b)}{\leq}-2\eta_t\frac{C}{G}\mathbb{E}\left[\lambda\norm{\mathbf{w}^{(t)}-\mathbf{w}^*}^2\right] \label{ineq_3rd_term_bound} 
	\end{align}
	where $(a)$ follows by \eqref{ineq_convecity1} and $(b)$ follows by \eqref{ineq_convecity2}. Using \eqref{ineq_2nd_term_bound} and \eqref{ineq_3rd_term_bound}, we can bound the expected value of \eqref{eq_se_bound} as
	\begin{align}
	&\mathbb{E}\left[\norm{\mathbf{w}^{(t+1)}-\mathbf{w}^*}^2\right]\nonumber\\
	&\leq\!\mathbb{E}\!\left[\norm{\mathbf{w}^{(t)}-\mathbf{w}^*}^2\right] \!+\! \eta_t^2C^2(1+d\sigma^2)\nonumber\\
	&\qquad\qquad-2\eta_t\frac{C}{G}\lambda\mathbb{E}\left[\norm{\mathbf{w}^{(t)}-\mathbf{w}^*}^2\right]\nonumber\\
	&=\!(1\!-\!2\eta_t\frac{C}{G}\lambda)\mathbb{E}\!\left[\!\norm{\mathbf{w}^{(t)}\!-\!\mathbf{w}^*}^2\right]\!+\!\eta_t^2C^2(1\!+\!d\sigma^2).
	\end{align} 
	Assume a learning rate of $\eta_t=\frac{G}{C\lambda t}$, so we have
	\begin{align}
	&\mathbb{E}\left[\norm{\mathbf{w}^{(t+1)}-\mathbf{w}^*}^2\right]\nonumber\\
	&\leq(1-\frac{2}{t})\mathbb{E}\left[\norm{\mathbf{w}^{(t)}-\mathbf{w}^*}^2\right]+\frac{G^2(1+d\sigma^2)}{\lambda^2t^2}\label{ineq_mse_rec}.
	\end{align}
	One can infer the following explicit bound for each term for some $a_1>0$ from the above recurrence relation:
	\begin{equation}
	\mathbb{E}\left[\norm{\mathbf{w}^{(t)}-\mathbf{w}^*}^2\right] \leq a_1\frac{G^2}{\lambda^2t}\label{ineq_mse_ex}.
	\end{equation} 
	According to \cite[Lemma 2]{rakhlin2011making}, the first term $\mathbb{E}\left[\norm{\mathbf{w}^{(1)}-\mathbf{w}^*}^2\right]$ can be bounded as $\mathbb{E}\left[\norm{\mathbf{w}^{(1)}-\mathbf{w}^*}^2\right] \leq \frac{4G^2}{\lambda^2}$.
	Thus, it is necessary to have $a_1\geq 4$. The next term with $t=2$ can be bounded by using the recurrence relation and the non-negativity of the mean square error (MSE) as 
	\begin{align}
	&\mathbb{E}\left[\norm{\mathbf{w}^{(2)}-\mathbf{w}^*}^2\right]\leq-\mathbb{E}\left[\norm{\mathbf{w}^{(1)}-\mathbf{w}^*}^2\right]+\frac{G^2(1+d\sigma^2)}{\lambda^2}\nonumber\\
	&\leq\frac{2G^2(1+d\sigma^2)}{2\lambda^2}.
	\end{align}
	To include this term, we should satisfy $a_1\geq 2(1+d\sigma^2)$. Similarly, the next term with $t=3$ can be bounded as
	\begin{align}
	&\mathbb{E}\left[\norm{\mathbf{w}^{(3)}-\mathbf{w}^*}^2\right]\leq\frac{G^2(1+d\sigma^2)}{4\lambda^2}\nonumber\\
	&\;\leq\frac{\frac{3}{4}G^2(1+d\sigma^2)}{3\lambda^2}.
	\end{align}
	Thus, we should satisfy also $a_1\geq\frac{3}{4}(1+d\sigma^2)$. For all terms with $t\geq 3$ the coefficient $(1-2/t)$ is always positive, so the bound can be shown by mathematical induction. Assume that \eqref{ineq_mse_ex} is true for $t=\tau\geq3$. By the recurrence formula, we can show that \eqref{ineq_mse_ex} also holds for $t=\tau+1$ as follows.
	\begin{align} 
	\mathbb{E}&\left[\norm{\mathbf{w}^{(\tau+1)}-\mathbf{w}^*}^2\right] \leq (1-\frac{2}{\tau}) a_1\frac{G^2}{\lambda^2\tau}+\frac{G^2(1+d\sigma^2)}{\lambda^2\tau^2}\nonumber\\
	&=a_1\frac{G^2}{\lambda^2(\tau+1)}\frac{1}{\tau^2}((\tau+1)(\tau-2)+\frac{\tau+1}{a_1}(1+d\sigma^2)) \nonumber \\
	&\!=\!a_1\frac{G^2}{\lambda^2(\tau+1)}(1\!-\!\frac{1}{\tau}(1-\frac{1+d\sigma^2}{a_1})\!-\!\frac{1}{\tau^2}(2-\frac{1+d\sigma^2}{a_1})) \nonumber\\
	&\overset{(a)}{\leq} a_1\frac{G^2}{\lambda^2(\tau+1)}
	\end{align}
	where $(a)$ holds if  $a_1 > 1+d\sigma^2$. Thus, \eqref{ineq_mse_ex} holds for every time instance if $a_1 \geq \max\{4, 2(1+d\sigma^2)\}$, i.e., we have
	\begin{equation}
	\mathbb{E}\left[\norm{\mathbf{w}^{(t)}-\mathbf{w}^*}^2 \right]\leq \max\{2, 1+d\sigma^2\}\frac{ 2G^2}{\lambda^2t}.
	\end{equation}
	Using this bound, a lower bound of the convergence rate can be obtained with the $\mu$-smoothness condition \eqref{ineq_smooth} as follows:
	\begin{align}
	&\mathbb{E}[\mathcal{L}(\mathbf{w}^{(t)})-\mathcal{L}(\mathbf{w}^*)] \leq \dfrac{\mu}{2}\mathbb{E}\left[\norm{\mathbf{w}^{(t)}-\mathbf{w}^*}^2 \right]\\
	&\leq\max\{2, 1+d\sigma^2\}\frac{\mu G^2}{\lambda^2t}.
	\end{align}
	The lower bound of \eqref{eq_utility} is obtained by taking the inverse of the convergence rate bound in each case.
	
	Consider next the transmission rate bound, for which we denote the noisy gradient as a sum of the clipped gradient and noise to exploit their boundedness and statistical properties, respectively, as follows:
	\begin{align}
	&h(\tilde{\mathbf{g}}_k^{(t)}) = h(\bar{\mathbf{g}}_k^{(t)}+\mathbf{Z}_k) \leq h(\bar{\mathbf{g}}_k^{(t)})+h(\mathbf{Z}_k)
	\label{eq_sum_diff_entrp}
	\end{align}         
	such that
	\begin{align}
	h(\bar{\mathbf{g}}_k^{(t)}) \leq& \dfrac{1}{2}\log\det(2\pi e\mathbf{K})\!=\! \dfrac{1}{2}\log\Big((2\pi e)^d\det(\mathbf{K})\Big) \label{eq_diff_ent}
	\end{align}
	where $\mathbf{K}$ is the covariance matrix of $\bar{\mathbf{g}}_k^{(t)}$ and $\det(\cdot)$ represents the determinant of a matrix. Denote the element of $\mathbf{K}$ in $i$-th row and $j$-th column as $K_{i,j}$ for all $i,j=1,2,\ldots,d$. The determinant of the covariance matrix can be bounded as 
	\begin{align}
	\det(\mathbf{K})&\overset{(a)}{\leq}\prod_{i=1}^dK_{i,i}=\prod_{i=1}^d\text{Var}\big(\bar{g}_{k,i}^{(t)}\big)\leq \prod_{i=1}^d\mathbb{E}\big[(\bar{g}_{k,i}^{(t)})^2\big] \label{eq_product_g2}
	\end{align}
	where $(a)$ follows from the Hadamard's inequality since a covariance matrix is positive-semidefinite. We can further bound $\det(\mathbf{K})$ by using the boundedness of the $\ell_2$-norm of the clipped gradient, i.e., $\sum_{i=1}^d\mathbb{E}[(\bar{g}_{k,i}^{(t)})^2]=\mathbb{E}\big[\sum_{i=1}^d(\bar{g}_{k,i}^{(t)})^2\big]\leq\mathbb{E}[C^2]=C^2$. Using the inequality of arithmetic and geometric means, the right hand side of \eqref{eq_product_g2} can be bounded as
	\begin{align}
	\prod_{i=1}^d\mathbb{E}\big[(\bar{g}_{k,i}^{(t)})^2\big] &\leq \left(\frac{1}{d}\sum_{i=1}^d\mathbb{E}\big[(\bar{g}_{k,i}^{(t)})^2\big]\right)^d \leq \left(\frac{C^2}{d}\right)^d.\label{eq:bound on determinantK}
	\end{align}
	Combining \eqref{eq_diff_ent}-(\ref{eq:bound on determinantK}), we obtain
	\begin{align}
	h(\bar{\mathbf{g}}_k^{(t)})\leq\dfrac{d}{2}\log\Big(\frac{2\pi eC^2}{d}\Big). \label{ineq_grad_diff_ent}
	\end{align}
	Furthermore, the differential entropy of the noise $\mathbf{Z}_k\sim\mathcal{N}(\mathbf{0},C^2\sigma_k^2\mathbf{I})$ is
	\begin{align}
	h(\mathbf{Z}_k)=\dfrac{1}{2}\log\det(2\pi eC^2\sigma_k^2\mathbf{I}_d) =\dfrac{d}{2}\log(2\pi eC^2\sigma_k^2). \label{eq_gauss_diff_ent}
	\end{align}
	Combining \eqref{eq_sum_diff_entrp}, \eqref{ineq_grad_diff_ent}, and \eqref{eq_gauss_diff_ent}, the proof of \eqref{eq_tr_rate_bound} follows. 
\end{appendices}

\end{document}